\documentclass[preprint,12pt]{elsarticle}

\usepackage{amssymb}
\usepackage{amsmath}	% 分段函数
\usepackage{mathrsfs}
\usepackage{bbm}
\usepackage{amssymb}
\usepackage{caption}
\usepackage{graphicx}
\usepackage{diagbox} % 斜线表格
\usepackage{subfigure} % 子图
\usepackage{stfloats}
\usepackage{array}
\usepackage{multirow}
\usepackage{url}

\journal{Journal}

\begin{document}

\begin{frontmatter}



\title{Perturb More, Trap More: Understanding Behaviors of Graph Neural Networks}


\author[a,b]{Chaojie Ji\corref{equ1}}
\ead{cj.ji@siat.ac.cn}
\author[a,b]{Ruxin Wang\corref{equ1}}
\ead{rx.wang@siat.ac.cn}
\author[a,b]{Hongyan Wu\corref{cor1}}
\ead{hy.wu@siat.ac.cn}

\cortext[equ1]{Equal contribution.}
\cortext[cor1]{Corresponding authors at: Shenzhen Institutes of Advanced Technology, Chinese Academy of Sciences, Shenzhen, China and Joint Engineering Research Center for Health Big Data Intelligent Analysis Technology.}
\address[a]{
Shenzhen Institutes of Advanced Technology, Chinese Academy of Sciences, Shenzhen, China
}
\address[b]{
Joint Engineering Research Center for Health Big Data Intelligent Analysis Technology
}

\begin{abstract}
While graph neural networks (GNNs) have shown a great potential in various tasks on graph, the lack of transparency has hindered understanding how GNNs arrived at its predictions. Although few explainers for GNNs are explored, the consideration of local fidelity, indicating how the model behaves around an instance should be predicted, is neglected. In this paper, we first propose a novel post-hoc framework based on local fidelity for any trained GNNs - \textbf{TraP2}, which can generate a high-fidelity explanation. Considering that both relevant graph structure and important features inside each node need to be highlighted, a three-layer architecture in TraP2 is designed: i) interpretation domain are defined by \textbf{T}ranslation layer in advance;
ii) local predictive behavior of GNNs being explained are probed and monitored by \textbf{P}erturbation layer, in which multiple perturbations for graph structure and feature-level are conducted in interpretation domain; iii) high faithful explanations are generated by fitting the local decision boundary through \textbf{P}araphrase layer. Finally, TraP2 is evaluated on six benchmark datasets based on five desired attributions: accuracy, fidelity, decisiveness, insight and inspiration, which achieves $10.2\%$ higher explanation accuracy than the state-of-the-art methods.
\end{abstract}


% \begin{highlights}
% \item Research highlight 0
% \end{highlights}


\end{frontmatter}


\section{Introduction}

The development of deep neural networks has shown outstanding performances in many domains \cite{badrinarayanan2017segnet, zhao2019object, alshemali2020improving}, which is partially attributed to the effectiveness of mining latent representations from Euclidean domain. By using regular grid convolution, convolutional neural networks (CNNs) effectively extract important semantic information from Euclidean data. However, there is an increasing number of applications which data are represented as graphs. Recently, Graph neural networks (GNNs) \cite{hamilton2017inductive, kipf2016semi, chami2019hyperbolic, ji2020hopgat} have been proposed and achieved breakthroughs in processing various graph structure data, such as point clouds, social network, chemistry molecules solved some classical field problems such as node classification \cite{abu2018n} and graph classification \cite{al2019ddgk}. With the advent of more high-precision models, requirements for a better understanding of the inner workings of a model are getting more attention. It is important to know the reason why this model produces such predictions. From the aspect of application, models need to provide transparent and high-precision solutions especially for some key scenarios, such as security, economy and healthcare. Meanwhile, interpretable approach can provide insight for improving models, and show underlying rules that are overlooked. Therefore, higher demands on the performances are raised for explainer: 1) the explainers need provide accurate, reliable and stable explanation for original model. 2) excellent interpretable approaches are expected to provide insight for model cognition and reveal potential rules that are neglected. However, the design of end-to-end network architecture reduces the transparency of the model, which hinders people to comprehend them.

To improve the transparency of deep neural networks, many researches of explanation technologies have been introduced and applied in recent years \cite{fong2017interpretable, niu2019pathological, lakkaraju2017interpretable, wagner2019interpretable}. Regrettably, existing explanation methods encountered barriers when they are applied to GNNs due to the particularity of graph structure. In recent two years, few excellent explanation approaches for GNNs have been studied. A qualified interpretation model should provide accurate explanation and be faithful to original model. Although learning a completely faithful explanation is usually impossible, a meaningful interpretation in the vicinity of prediction being explained is possible \cite{ribeiro2016should}. Unfortunately, taking node classification as an example, prior researches only use the GNN's prediction behavior information of node being explained to generate explanations, which neglects the other local behaviors near the decision boundary of the node. These behavior informations are very useful for providing a local faithful explanation. To our best knowledge, there is no method considering the ``vicinity" in graph-structured data for a post-hoc explanation based on local fidelity at present. In addition, each feature component is node-dependent that each feature can make different contribution for distinct nodes and the prediction behaviors of GNNs. It is crucial for evaluating and integrating them for an accurate and reliable explanation. However, existing methods ignore this. For example, the GNNExplainer \cite{ying2019gnnexplainer} only coarsely provides a shared score for the feature components of all nodes, which leads to a sub-optimal interpretation effect.

To address above problems, we propose a novel post-hoc and model-agnostic explanation framework for GNNs, which provides a ``broad'' insight of original model (Figure \ref{fig:overview}). To achieve this, the proposed approach includes a three-layer architecture design which is named \textbf{TraP2}: i) \textbf{Tra}nslation Layer is adopted to realize the transformation from the original problem to interpretation domain according to different tasks. ii) To ``trap'' richer predictive behaviors near a local decision boundary of the object being explained, the \textbf{P}erturbation Layer probes and monitors local behaviors by specially designed strategy from both graph structure-level and feature-level. Meanwhile, a novel perturbation energy level is proposed for measuring the obtained attention of each perturbation. iii) Based on perturbed instances, \textbf{P}araphrase Layer finds high-faithful explanations by fitting the local decision boundary, which also provides insight in both graph structure-level and node-dependent feature-level. Finally, to verify the effectiveness of the presented model, we evaluate it on multiple datasets \cite{ying2019gnnexplainer} for node and graph classification, and compare our results with other state-of-the-art methods. The results demonstrate the effectiveness of our work from accuracy, fidelity and contrastivity, which outperforms other approaches for all tasks. In summary, the main contributions of this paper can be summarized as follows:

\begin{itemize}
	\item We first put forward a novel post-hoc framework \textbf{TraP2} based on local fidelity of any GNN models for different recognition tasks, which generates high-fidelity explanations.

	\item Novel perturbation strategy and perturbation effect estimation method designed for graph data are proposed. Furthermore, our model provides a more fine-grained explanation on node-dependent feature-level than prior works.

	\item Compared with the state-of-the-art GNN explanation approaches, the proposed method achieves the top performance on multiple benchmark datasets. %Especially, achieves $13.3\%$ higher explanation accuracy than the state-of-the-art method.
\end{itemize}

The remainder of this paper is organized as follows: Section 2 is the related works containing the introduction of the graph neural networks, current non-graph and graph neural networks-based interpretability methods. In Section 3, we introduce the details of our proposed explanation approach. Experimental results for node and graph classification tasks are given in Section 4. Finally, Section 5 concludes this paper and prospects some future works.

\section{Related Work}

\subsection{Graph neural networks}

In recent years, graph neural networks have been successfully applied to a wide variety of fields such as computer vision \cite{chen2019multi, shi2019two}, natural language processing \cite{yao2019graph}, recommender systems \cite{ying2018graph} and healthcare \cite{shang2019gamenet}. GNNs effectively handle the complex relationship between objects in the graph structure following a neighborhood aggregation scheme, where the features of a node is computed by recursively assembling and transforming features of its local neighbors \cite{zhang2020deep, wu2020comprehensive}. They can be divided into two categories: spectral \cite{li2018adaptive, zhuang2018dual} and spatial methods \cite{xu2018powerful, zhang2018end, chiang2019cluster}. The spectral method is defined via graph Fourier transform and convolution theorem, which takes the graph Laplace matrix as an important tool and does not explicitly use the information propagation mechanism on the graph. Bruna et al. \cite{bruna2013spectral} proposed a method to conduct convolution in the spectral domain adopting the Fourier basis of a given graph. Levie et al. \cite{levie2018cayleynets} introduced a new spectral domain convolutional architecture using a new class of parametric rational complex functions (Cayley polynomials) that can specialize on frequency bands of interest. On the other hand, spatial approaches define convolution operations in the vertex domain, operating on spatially local neighborhood nodes. For instance, Ying et al. \cite{ying2018hierarchical} proposed a differentiable graph pooling method for GNNs, which can be used to obtain a representation of an entire graph by summing the features of all nodes in the graph. Veli\v{c}kovi\'{c} et al. \cite{velivckovic2017graph} presented graph attention networks operated on graph-structured data, assigning different weights to different nodes within a neighborhood.

\subsection{Non-graph neural networks interpretability methods}

Many well-studied explanation techniques employ gradient-based backpropagation to calculate saliency maps for original input. Some prominent methods of this category include Class Activation Mapping (CAM) \cite{zhou2016learning}, Gradient-weighted Class Activation Mapping (Grad-CAM) \cite{selvaraju2017grad}, (Grad-CAM++) \cite{chattopadhay2018grad}, Excitation Back-Propagation (EB) \cite{zhang2018top} and Layer-wise Relevance Propagation (LRP) \cite{bach2015pixel}. The CAM and its generalization Grad-CAM measure the linear combination of each layer's activations and class-specific weights or gradients of original model. The EB improves gradient maps by introducing contrastive top-down attention. And the LRP adopts layer-wise relevance propagation to achieve a pixel-wise decomposition. The above approaches introduce different backpropagation heuristics, which can focus on salient notions of input data. However, they are not model-agnostic, with most of them being limited to original network framework and/or many necessary modifications of model structure \cite{mahendran2016salient}. LIME \cite{ribeiro2016should} is a representative system of model-agnostic approaches, which adopts the self-explanatory linear regression to local area and pinpoints important features based on the regression coefficients. The SHAP model \cite{lundberg2017unified} uses the Shapley values of a conditional expectation function of the original model to measure the importance of each feature for a particular prediction. %Also, other extended works with complex linear models based LIME (e.g., decision tree \cite{bastani2017interpreting}) are proposed and achieve better performance.

\subsection{Graph neural networks interpretability methods}

To the best of our knowledge, few explainers for GNNs are explored recently. Pope et al. \cite{pope2019explainability} extended the gradient-based saliency map methods to GNNs, which utilizes the network parameters and classifier output of original GNNs to construct the activation response of the corresponding neurons. Baldassarre et al. \cite{baldassarre2019explainability} employed two main classes of techniques, gradient-based and decomposition-based, to learn important components of input that also relies on propagating gradients/relevance from the output to the input of original model. Ying et al. \cite{ying2019gnnexplainer} proposed an explanation method by identifying a small subgraph structure and a subset of node features that maximizes the mutual information with original GNNs prediction in entire input graph. Although the above works have made breakthrough progress in GNNs interpretation, the consideration of local fidelity and fine-grained explanation on feature-level is neglected, which can provide abundant information to explain original GNNs' behaviors.

\begin{figure*}[!t]
	%\centering
	\centerline{\includegraphics[width=1\textwidth]{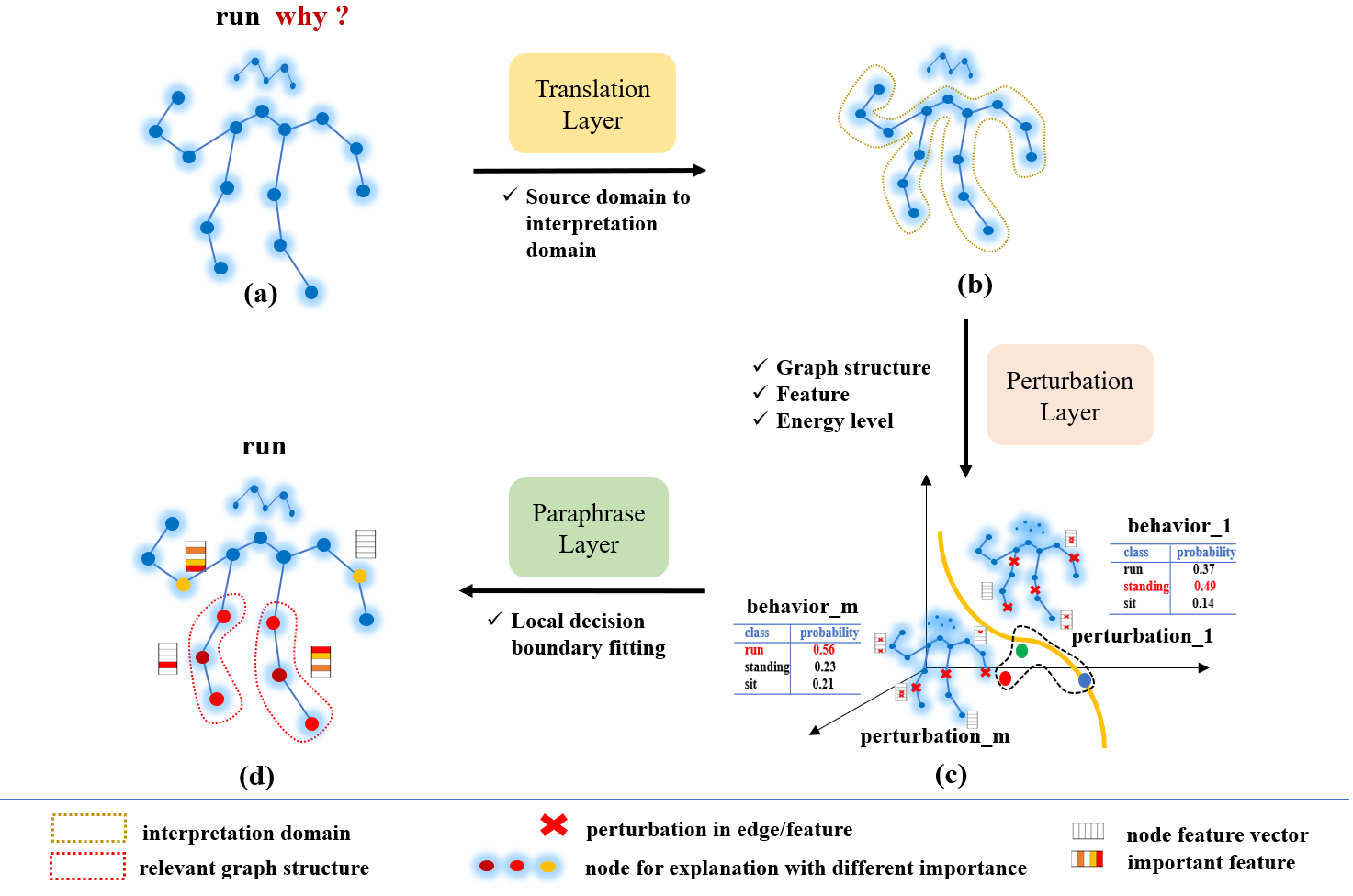}}
	\caption{Overview of the proposed TraP2 framework, which contains a three-layer architecture with \textbf{Tra}nslation layer, \textbf{P}erturbation layer and \textbf{P}araphrase layer.}
	\label{fig:overview}
\end{figure*}

\section{Preliminary}
\subsection{Formulaic Definition of Explanation on Graph}
% Definition of Graph-structured Data
A graph can be formulated as $G = (V, E, X)$, in which $V$ is a set of $n$ nodes $\{ v_1, ..., v_n\}$, $E$ denotes a set of edges and $X = \{x_1, ..., x_n | \forall x_i\in \mathbb R^d\}$ represents the attributes for all nodes. $d$ is the number of features. For simplicity, we assume all edges have an identical type, thus an adjacency matrix $A \in \mathbb R^{n \times n}$ can be assigned to represent $V$ and $E$.
% Definition of Graph Tasks, Node and Graph Classification
$Y \in \{1, ..., C\}$ refers to the node or graph label according to different tasks.

Whatever node or graph classification, an explanation can be uniformly considered as a small subgraph $\hat G = (\hat V, \hat E, \hat X)$ which makes a great contribution for the prediction of the GNNs being explained. The combination of $\hat V \subseteq V $ and $\hat E \subseteq E$ highlights the explanation graph structure. Furthermore, the interpretable features of nodes inside $\hat V$ are marked as $\hat X$.

For simplifying the following formulation, we explicitly define two functions: generating reachability matrix and fetching specified elements from an adjacency matrix.
Reachability matrix records the connectivity between nodes within $k$-hop. It is able to be derived from multiplying the adjacency matrix $A$ by itself. We formulate the function of generating it as $[A^k]$. Fetching an element of the $i$th row and $j$th column from a given matrix $M$ is formulated as $[M]_{i, j}$. In particular, $[M]_{i, :}$ represents fetching $i$th row in $M$.

\subsection{Graph Neural Networks}
% Definition of GNN
GNNs update the representation of each node through summarizing the local information similar to convolution operation in CNNs. At each layer $l$, graph structure $(V, E)$ is kept unchanged, only the features of the graph are updated:

\begin{equation}
\label{Preliminary:GraphNeuralNetworks:E1}
m^{l+1}_{i} = \sum_{j\in \mathcal N_i} \mathcal F(h^l_i, h^l_j, e_{ij})\ ,\ h^{l+1}_i = \mathcal G(m^{l+1}_i, h_i^l)
\end{equation}
where $\mathcal N_i$ defines the neighbor nodes around node $i$. $\mathcal F$  and $\mathcal G$ are message and update function respectively. The representation of node $i$ in layer $l$ is denoted as $h^l_i$. $h^0_i$ is initialized as $x_i$. $e_{ij}$ symbolizes the type of edge from node $i$ to $j$.
The updated representation after final layer $L$ can be mapped as $f: h^L \Rightarrow Y$ for a specific task, i.e. node classification (regression), graph classification (regression) and etc.. Formally, GNNs can be formulated as $f(A, X)$.

% In node classification, GNNs are used to model $p(y|x_V)$ for each labeled nodes. GNNs predict the label in following:
% $$
% p(y|X_V) = softmax(Wh_n)
% h=GNN(x_V, E)
% $$
% where $h$ is the learnt representation of all nodes from GNNs. $GNN$ iteratively aggregate $h$ according to the representation of current value $h_i$ and neighbors $h_j^{l-1}, i \in h_{NB(n)}^{l-1}$. Then the aggregated value is updated in different ways, e.g. graph convolutional layer, graph attention layer and neural message passing layer. Once pre-defined number of layers reach, the final $h$ is fed into softmax classification.

\section{Method}
In this section, we describe the uniform framework of our model - TraP2. TraP2 has three components: 1) Translation Layer: accessible subgraphs to be explained are remained from complete input graphs; 2) Perturbation Layer: translated subgraphs are respectively perturbed in aspects of graph structure and node feature. Meanwhile, the degree of these perturbation is assessed as ``perturbation energy level''; 3) Paraphrase Layer: a local faithful interpretation is trained to explain the behaviors of GNNs, which is shown in Figure \ref{fig:overview}.

In this section, we first introduce the explanation process of node classification on node $i$ and then extend it into other graph tasks.

\subsection{Translation Layer: Transference from Source Domain to Interpretation Domain}
\label{Translation_Layer}
In some cases, the domain to be explained for one node is not the entire graph (Source Domain) as shown in Figure \ref{fig:overview} (a). To be more precise, GNNs with $k$ layers can only aggregate messages from $k$-hop, e.g. 2-3, neighbors \cite{kipf2016semi}.
According to Equation (\ref{Preliminary:GraphNeuralNetworks:E1}), final representation of node $i$, $h^l_i$, is obtained in $l$ recursive updates. In each update, nodes located in one-hop farther from node $i$ are aggregated. Thus only nodes within $l$-hop from node $i$ is considered in GNNs. It directly results in that the feasible region for explanation is merely a small subgraph (Interpretation Domain) as shown in Figure \ref{fig:overview} (b) instead of entire graph Figure \ref{fig:overview} (a). Benefited from it, the solution space can be potentially shrank.

Inspired by it, translation layer is applied to transforms the source graph-structured domain into a limited interpretable domain for node $i$:
\begin{equation}
\label{Method:TranslationLayer:E1}
V^\mathcal I = \{x| x>0, \forall x \in [A^k]_{i,:}\}
\end{equation}

\begin{equation}
\label{Method:TranslationLayer:E2}
A^\mathcal I = [A]_{V^\mathcal I\times V^\mathcal I} \in \{0, 1\}^{\hat n \times \hat n}
\end{equation}
where $V^\mathcal I$ and $A^\mathcal I$ denote the node set and corresponding adjacency matrix in interpretation domain respectively. $\hat n$ is the number of elements in set $V^\mathcal I$.

\subsection{Perturbation Layer: Turbulence in Interpretation Domain}
\label{Perturbation Layer}
% three points for main three components
% Although learning a completely faithful explanation is impossible, a meaningful interpretation in the vicinity of prediction being explained is possible. Unfortunately, to our best knowledge, there is no method considering the ``vicinity" in graph-structured data and monitors on the local behaves on this local areas for an explanation.
In order to realize a local-fidelity based explanation, the strategies of perturbation are proposed to probe the responses of GNNs in local vicinity.
Accordingly, a series of novel perturbation patterns are designed for graph. These perturbed instances can be distinguished from the aspects of both graph structure and feature as shown in Figure \ref{fig:overview} (c).
In particular, we introduce the concept of perturbation energy level that magnitude of each disturbance is quantified. And it'll be further delivered into paraphrase layer for establishing attention for each perturbation. Finally we monitor and record the corresponding behavior response of GNNs to be explained in the disturbances.

\subsubsection{Perturbation on Graph Structure}
We first define an action set on graph structure: i) adding new edges between nodes in $V^\mathcal I$, ii) removing existing edges from $A^\mathcal I$.

A random action variable, $z^E$, is applied to trigger these two actions alternatively:
\begin{equation}
\label{Method:PerturbationLayer:E1}
z^E \sim Bernoulli(1, p_1)
\end{equation}
where $p_1$ is the probability for actions.
As a graph can be represented as a binary adjacency matrix which can be deformed through alternating $0$ and $1$ for each element in it.
The composition of a perturbed graph $A^P$ under different patterns can be decomposed as:
% \begin{equation}
% \label{Method:PerturbationLayer:E3}
% [M^A]_{j,k} = |\tau (z^E-1, [S^A]_{j,k})|
% \end{equation}
\begin{equation}
\label{Method:PerturbationLayer:E2}
[A^\mathcal P]_{j,k} = \left\{
\begin{aligned}
{[A^\mathcal I]}_{j,k} \oplus (z^E-1), & & Adding\ Pattern\\
{[A^\mathcal I]}_{j,k} z^E, & & Removing\ Pattern\\
{[A^\mathcal I]}_{j,k} \oplus z^E, & & Adding\ \&\ Removing\\
\end{aligned}
\right.
\end{equation}
in which $[A^\mathcal P]_{j,k} \in \{0, 1\}$ and $\oplus$ denotes \emph{xor} operation. Actions of adding and removing edges are alternatively or simultaneously regarded in these three patterns respectively.

% The selection of function $\tau(\cdot)$ determines different perturbation patterns. Adding and removing edges are simultaneously considered in addition operation; multiplication operation only apply removing links; Xor operation merely employs adding connections.

Furthermore, specific constraint on perturbation patterns can be additionally imposed. For an instance, if $A^\mathcal I$ is relatively sparse that number of nodes located within 1-hop from node $i$ is extremely smaller than farther hops, an arbitrary perturbation on $A^\mathcal I$ probably results in a great impact - far from the vicinity.
% Account for review 2 - 1
Given a specified case that all directly connected node around node $i$ is a single vertex $j$ which is further expended with many other farther nodes, it implies that node $j$ provide a most valuable clue for analyzing node $i$ due to the smallest hop distance. In such situation, large indiscriminate perturbations probably result in constant absences of node $j$ and thus explainer is forced to neglect this informative node in paraphrase layer. To solve it, the perturbations occurred on the edges that directly connect with node $i$ should be prohibited as $[A^\mathcal P]_{i,j} = [A^\mathcal I]_{i,j}, \forall (i,j) \in E$.

\subsubsection{Perturbation on Feature}
Similarly, we design a perturbation pattern for feature by scaling or masking its representation. $z^F$ is applied as the random action variable for perturbation on feature:
\begin{equation}
\label{Method:PerturbationLayer:E4}
z^F \sim \left\{
\begin{aligned}
Bernoulli(1, p_2), & & Masking\ Pattern\\
\mathcal N(0, 1), & & Scaling\ Pattern\\
\end{aligned}
\right.
\end{equation}
Features of node $i$ and $z^F$ are combined as:
\begin{equation}
\label{Method:PerturbationLayer:E5}
[X^\mathcal P]_{i,d} = [X]_{i,d} z^F
\end{equation}

\subsubsection{Perturbation Energy Level}
``Perturbation energy level" explicitly quantifies the energy consumption of perturbation according to both graph structures and features.
% Account for review 3 - 5
Obviously, more complicated perturbations consume huger amount of energy, which always introduce more severe deformation that more edges are removed and added and features inside nodes are masked. Then corresponding response from GNNs also quite differs from the original graph. In contrast, samples produced by slighter perturbations with smaller energy consumption always locate in an immediate vicinity of the original samples.
As a result, observation on smaller perturbations with slightly altered prediction provides a more important clue to track the ``logic" of GNNs, emphasized in learning stage of paraphrase layer.

For the respect of graph structure, we assume that the combination of the distance between perturbed position and node $i$, and the deformation degree - scale of altered edges - jointly indicate the energy level. Concretely, the closer perturbation and larger deformation usually consume more energy and vice versa.

We define the measurement of perturbation distance as hop value and formulate a normalized coefficient for $k$-hop as:
\begin{equation}
\label{Method:PerturbationEnergyLevel:E1}
w_k = \frac{K}{k+1}
\end{equation}
\begin{equation}
\label{Method:PerturbationEnergyLevel:E2}
\alpha_k = \frac{e^{(w_k)}}{\sum_{i\in \{1, ..., K\}}e^{(w_i)}}
\end{equation}
where $K$ is the pre-defined maximum, e.g. number of layers in original GNNs. Finally, we obtain the energy level by combining the coefficients with the deformation:
\begin{equation}
\label{Method:PerturbationEnergyLevel:E4}
\gamma_A = \sum_{k=1}^K \alpha_k sim([(A^\mathcal P)^k]_{i,:}, [(A^\mathcal I)^k]_{i,:})
\end{equation}

For the energy level of feature, we measure the similarity between original and perturbed node feature:

\begin{equation}
\label{Method:PerturbationEnergyLevel:E5}
\gamma_X = \sum_{i=1}^N sim([X^\mathcal P]_{i,:}, [X]_{i,:})
\end{equation}
where
\begin{equation}
\label{Method:PerturbationEnergyLevel:E3}
sim(u, v) = e^{(\frac{-cosine(u, v)^2}{\delta^2})}
\end{equation}
in which $\delta$ is the width with distance function $cosine(\cdot)$. It suggests that higher energy is consumed with more difference in node features.

To this end, complete energy level is defined as follows:
\begin{equation}
\label{Method:PerturbationEnergyLevel:E6}
\gamma = \lambda_A \gamma_A  + \lambda_X\gamma_X
\end{equation}
where $\lambda_A$ and $\lambda_X$ control the balance between perturbation on graph structure and feature.

\subsubsection{Monitor on Multiple Perturbations}
Given a single instance, the complex decision boundaries of GNNs being explained can be hardly identified by limited witnesses. Thus multiple perturbations are applied and we formulate a series of independent instances perturbed as $\{(A^\mathcal P_{(j)}, X^\mathcal P_{(j)}, \gamma_{(j)})\}_{j=1}^m$ in which $m$ is the frequency of the perturbations. Meanwhile, for each perturbation, we constantly monitor and record the behavior feedback  $f(y|A^\mathcal P_{(*)},X^\mathcal P_{(*)})$ of the GNNs being explained. $*$ represents a perturbation.

\subsection{Paraphrase Layer: Explanation on Interpretation Domain}
Once multiple behavior responses of GNNs under various perturbations are collected and learnt sequently, a local decision boundary can be identified by the explainer.

\subsubsection{Learning Phase}
% Account for review 2 - 2
In order to learn a local faithful explanation, we denote an explainer as $g$ which can be any kind of potentially explainable models, including linear models and non-linear models. $w_\theta \in \mathbb R^{\hat n d}$ is the trainable parameters, the size of which is completely correlated with the scale of $A^\mathcal I_{(*)}$ and $X$. $d$ is unchangeable and derived from the number of features inside $X$. As mentioned in section \ref{Translation_Layer}, the size of $A^\mathcal P_{(*)}$ is largely reduced from $A$ according to a transformation executed by the translation layer. That is to say that $\hat n$ equals to the number of vertices within limited hop - defined by the GNNs to be explained - from the explained node. Accordingly, it ensures that $\hat n \ll n$. And $g$ has ability to generate explanations for an even relatively large graph.

The explainer $g$ is calculated as:
\begin{equation}
\label{Method:ParaphraseLayer:E1}
g(y|A^\mathcal P_{(*)}, X^\mathcal P_{(*)}; w_\theta)=\sigma(w_{\theta}({||}_{j=1}^n ([(A^\mathcal P_{(*)})^k]_{i,j} \cdot [X^\mathcal P_{(*)}]_{j,:})))
\end{equation}
where symbol $||$ indicates a concatenation operation, $\cdot$ is a scalar multiplication and $\sigma$ denotes a nonlinearity function.

In addition, unlike GNNExplainer, the explanation of TraP2 on feature is node-dependent as shown in Figure \ref{fig:overview} (d).
% It effectively improve the explanation accuracy.

We combine explainer $g$ with GNNs being explained $f$ to fit a local decision boundary:
% \mathbbm{1}
\begin{align}
\label{Method:ParaphraseLayer:E2}
% \underset{\theta}{min}\sum_{j=1}^m \sum_{c=1}^C \mathcal{L} (\gamma_{(j)}, f(y=c|M^A_{(j)}, M^F_{(j)}), g_{\theta}(y=c|M^A_{(j)}, M^F_{(j)})) + \Omega(\theta)
%\begin{align}
% x &= \sqrt {1-y^2}\\
% x &= \sqrt[3]{1-y^3}
\underset{w_\theta}{min}\sum_{j=1}^m \sum_{c=1}^C \frac{1}{\gamma_{(j)}} \mathcal{L} (f(y=c|A^\mathcal P_{(j)}, X^\mathcal P_{(j)}), \nonumber\\
g(y=c|A^\mathcal P_{(j)}, X^\mathcal P_{(j)})) + \lambda\Omega(w_\theta)
%\end{align}
\end{align}
where $\mathcal L$ is a measurement of the difference between explainer and GNNs in the locality. Fewer energy consumption $\gamma_{(j)}$ causes more attention on this perturbation in that they are closer to the local decision boundary. $\Omega(w_\theta)$ is regarded as a regularization term to encourage $w_\theta$ to be discrete and be interpretable by humans.

\subsubsection{Identifying Phase}
In our work, the obtained parameters $w_\theta$ directly represent the contribution score of a single feature for a node. Further we compute the contribution (explanation) of node $j$ with $w_\theta$:

\begin{equation}
\label{Method:ParaphraseLayer:E3}
I^{i}_j = \sum_{q=j*d}^{(j+1)*d}|[w_{\theta}]_q|
\end{equation}
where $|[w_{\theta}]_q|$ indicates the contribution, an absolute value, from the $d$th feature inside node $j$.
% Account for review 2 - 3
A main goal that explainers aim at is locating nodes that relatively maximum extents of contributions are made. Therefore, the direction of contributions, either positive or negative, have no effect on quantifying the extent.

At last, by sorting of the contribution scores across $\{I^{i}_1, ..., I^{i}_{\hat n}\}$ and $\{|[w_{\theta}]_1|,$ $...,$ $|[w_{\theta}]_{\hat n d}|\}$, we can achieve the $\hat V \subseteq V^\mathcal I$ and $\hat X$ with higher scores respectively. Furthermore, $\hat E$ can be extracted from existing edges inside $E$ and nodes belonging to $\hat V$. Thus, a complete explanation of node $i$, $\hat G = (\hat V, \hat E, \hat X)$, is determined.

\subsection{Extension on Graph Classification Task}
Our proposed TraP2 can not only explain on node classification but also other graph machine learning tasks, e.g. graph classification.

In node tasks, only one node $i$ need to be explained, thus TraP2 appropriately executes one time. However, every node in the graph all makes contributions to graph prediction. According to it, for each node, we independently perform a translation, perturbation and paraphrase as mentioned above. Specifically, there are two small changes in paraphrase layer. i) The $c$ in Equation (\ref{Method:ParaphraseLayer:E2}) belongs to label of graph instead of node; ii) In identifying stage, the contribution scores of each node $j$ must be pooled across all nodes:

\begin{equation}
\label{Method:ParaphraseLayer:E4}
I_j = \frac{1}{n} \sum_{i=1}^{n} I_j^i
\end{equation}

% ii) link prediction
% Two nodes located at the end of the link being explained are interpreted through TraP2 respectively.

% Besides of below node classification, we also can easily apply our proposed 4p architecture into other tasks.
% For multiple instances, we similarly implement below process. Unless we no longer only select a single node to explain, we simultaneously select multiple nodes to project into disturbed space and train the loss function with all nodes.
%
% The explanation on graph classification is similar. We train model $g$ with all nodes inside the graph. Furthermore, we use the ground truth of graph instead of label of nodes to train model. The final trained $w_{\theta}$ are averaged across dimension to indicate import factors for every nodes.
%
% The explanation of link prediction sample the two nodes located in the ends of each possible connection. The process of explainable and locality projector keep unchange. However, in explanation phrase, we concatenation these two $M^E$ into one and attempt to simulate the ground truth, i.e. possibility of these ones' connection.

\section{Experiment}
In this section, we conduct experiments on two kinds of tasks, node and graph classification to evaluate the performance of TraP2.

\subsection{Datasets}
For these tasks, we follow existing study to apply the same benchmark datasets \cite{ying2019gnnexplainer}.
\begin{itemize}
  \item \textbf{BA-SHAPES}
  is a Barab\'{a}si-Albert (BA) graph with $300$ nodes and $80$ five-node house attachments which are randomly attached. The classes are determined by that nodes locate on the top, middle, bottom or out of a house.

  \item \textbf{BA-COMMUNITY}
  consists of two BA graphs as in Figure \ref{fig:node}. The definition of class for each graph is consistent with BA-SHAPES. In addition, normally distributed features are assigned for each node.

  \item \textbf{TREE-CYCLE}
  is a $8$-level balanced binary tree randomly attached with $80$ six-node cycles. The classes are distinguished by whether nodes locate on the cycles.

  \item \textbf{TREE-GRID}
  is similar with TREE-CYCLE except of that six-node cycles are replaced by $3$-by-$3$ grids.

  \item \textbf{MUTAG}
  is a dataset composed of $4,337$ molecule graphs which are labeled with the mutagenic effect on the Gram-negative bacterium S.typhimurium ~\cite{debnath1991structure}.

  \item \textbf{REDDIT-BINARY}
  is a dataset of $2,000$ graphs that online discussion threads on Reddit are recorded. Inside each graph, nodes indicate users and edges stand for the reply between users. The labels of the graphs are the types of user interactions \cite{yanardag2015deep}.
\end{itemize}

%
%   \item \textbf{Fidelity}
%   Fidelity was applied to measure that the difference in accuracy obtained by reserving only top n related nodes. It explicitly represents the ability of recovering the prediction of explained GNNs.
%
%   \item \textbf{Contrastivity}
%   Contrastiviy was designed to capture the intuition that important factor belongs to top n explanation should be larger than the unselected nodes. Larger contrastivity, more interpretable ability for explanation.
%
% \end{itemize}

\subsection{Baseline Methods}
We compare TraP2 with four baseline methods:

\begin{itemize}
  \item \textbf{Random}
  is an approach that all explanation elements are randomly generated.

  \item \textbf{Greedy}
  is a method that the nodes causing the highest accuracy difference from original GNNs are added iteratively, when the edges between these nodes and the node being explained are masked one by one.

  \item \textbf{Grad}
  is similar to a saliency map method that the gradient is derived from the loss of GNNs being explained \cite{pope2019explainability}.

  \item \textbf{GNNExplainer}
  explores a small subgraph of the entire graph to maximize the mutual information with the prediction of original GNNs.

\end{itemize}

% \multirow{2}*{BA-COMMUNITY} & \multirow{2}*{1.6} & \multirow{2}*{65.3} &	\multirow{2}*{66.4} & \multirow{2}*{68.6} & \multirow{2}*{\textbf{71.2}} \\
% ~\\

% \begin{table*}[!htbp]
%   \centering
%   \begin{tabular}{c|ccccc}
% 		\hline
% 		% \multicolumn{1}{c|}{DU Vector}& \multicolumn{6}{c}{Fine-grained Classification}\\
% 		Dataset & Random & Greedy & Grad & GNNExplainer & TraP2\\
%     \hline
%      BA-SHAPES & 1.9 & 60.8 & 81.4 & 81.7 & \textbf{81.9}  \\
%      BA-COMMUNITY & 1.6 & 65.3 &	66.4 & 68.6 & \textbf{71.2} \\
%      TREE-CYCLE & 47.7 & 56.1 & 71.7 & 71.8 & \textbf{72.2} \\
%      TREE-GRID & 74.6 & 75.5 & 70.2 & 76.3 & \textbf{88.0}\\
% 		\hline
%   \end{tabular}
%   \caption{Explanation Accuracy.}
%   \label{Accuracy}
% \end{table*}

\subsection{Implementation}
% Account for review 2 - 8
The implementation and hyperparameters \footnote{\url{https://github.com/RexYing/gnn-model-explainer}} for GNNs to be explained and GNNExplainer are completely derived from the original work.
For TraP2, in translation layer, we set $k$ in Equation (\ref{Method:TranslationLayer:E1}) to 3 corresponding to the trained GNNs. Sample rate $p_1$ and $p_2$ in Equation (\ref{Method:PerturbationLayer:E1}) and (\ref{Method:PerturbationLayer:E4}) are respectively set as 0.5 and 0.8 for all datasets. The width $\delta$ with distance function in Equation (\ref{Method:PerturbationEnergyLevel:E3}) is assigned as 25.
% Account for review 2 - 4
We choose 1 for both $\lambda_A$ and $\lambda_X$ in Equation (\ref{Method:PerturbationEnergyLevel:E6}).
The frequency of perturbation is 1500. L1 regularization is selected as our regularization term $\Omega(w_\theta)$ in Equation (\ref{Method:ParaphraseLayer:E2}). In paraphrase layer, $g$ is trained in 300 epochs with 0.01 learning rate.
% Account for review 2 - 1
To account for the property of TREE-CYCLE and TREE-GRID datasets that number of nodes within 1-hop is smaller than 2 and 3-hops, we constrict the perturbation on graph structure that edges belonging to 1-hop from the node being explained are kept unchanged as mentioned in Section \ref{Perturbation Layer}.

\subsection{Investigation on TraP2}
In our work, five desired attributions: accuracy, fidelity, decisiveness, insight and inspiration are investigated for evaluating the performance of all explainers.
\begin{table}[!htbp]
  \centering
  % \resizebox{\textwidth}{12mm}
  \begin{tabular}{c|ccccc}
		\hline
		% \multicolumn{1}{c|}{DU Vector}& \multicolumn{6}{c}{Fine-grained Classification}\\
		Dataset & Random & Greedy & Grad & GNNExplainer & TraP2\\
    \hline
     BA-SHAPES & 1.9 & 60.8 & 81.4 & 81.7 & \textbf{81.9}  \\
     \hline
     BA-COMMUNITY & 1.6 & 65.3 & 66.4 & 68.6 & \textbf{71.2} \\
     \hline
     TREE-CYCLE & 47.7 & 56.1 & 71.7 & 71.8 & \textbf{72.2} \\
     \hline
     TREE-GRID & 74.6 & 75.5 & 70.2 & 76.3 & \textbf{86.5}\\
		\hline
  \end{tabular}
  \caption{Explanation Accuracy on four benchmark datasets.}
  \label{Accuracy}
\end{table}
\begin{table}[!htbp]
  \centering
  \begin{tabular}{c|c|ccc}
    \hline
    Dataset&Metrics&Grad&GNNExplainer&TraP2\\
    \hline
    \multirow{2}*{BA-SHAPES} & Fidelity & 0.042 & 0.030 & \textbf{0.020}\\
    ~ & Contrast & 1.01 & 1.47 & \textbf{6.98}\\
    \hline
    \multirow{2}*{BA-COMMUNITY} & Fidelity & 0.15 & 0.11 & \textbf{0.09} \\
    ~ & Contrast & 1.06 & 1.51 & \textbf{5.08}\\
    \hline
    \multirow{2}*{TREE-CYCLE} & Fidelity & 0.275 & 0.406 & \textbf{0.272}\\
    ~ & Contrast & 1.00 & 1.06 & \textbf{3.51}\\
    \hline
    \multirow{2}*{TREE-GRID} & Fidelity & 0.718 & 0.069 & \textbf{0.057}\\
    ~ & Contrast & 1.01 & 1.45 & \textbf{4.50}\\
    \hline
  \end{tabular}
  \caption{Explanation Fidelity and Contrast on four benchmark datasets.}
  \label{Fidelity_Contrastivity}
\end{table}
\subsubsection{Question 1: Can TraP2 make an accurate explanation for ground-truth knowledge?}
One of the most essential criterion of an explanation is the accuracy that the interpretation should match ground truth knowledge exactly.
We define the ground truth explanation for BA-SHAPES, BA-COMMUNITY, TREE-CYCLE and TREE-GRID as their attachments, i.e. house, cycle and grid shape. For each node being explained, we train an explainer and sort the contribution scores of each node. We only preserve the top $n$ nodes - number of the nodes inside ground truth - with higher scores and calculate the matching accuracy rate with the ground truth. Intuitively, a better explanation is expected to  achieve a higher accuracy. The results of node classification on four datasets are shown in Table \ref{Accuracy}. TraP2 outperforms other baseline methods on all datasets, even up to $10.2\%$ higher than the state-of-the-art approach in TREE-GRID that most complicated structures of grid and tree are combined tightly. It is noteworthy that Random method surprisingly obtains a reasonable accuracy in TREE-CYCLE and TREE-GRID. The reason is that the scale of interpretation domain is relatively small.

\subsubsection{Question 2: Is TraP2 faithful to the authentic prediction?}
A faithful explainer is attributed as a repeater that the original behaviors of GNNs can be reproduced without loss. That is to say that either entire graph or explanation subgraph is fed into trained GNNs, the outputs ought to be identical.
% Account for review 2 - 6
However, there is no guarantee that ground truth explanations are consistency with the internal logic of GNNs explained. From another aspect, fed a subgraph composed of ground truth explanations, GNNs may produce predictions which is extremely different from original outputs. In a conclusion, explainers with high accuracy are not the equivalent of the interpreters with high fidelity.

We formally formulate the fidelity as the absolute difference between the prediction from the entire graph and the explanation subgraph consisting of the $n$ nodes from Question 1. Smaller difference explicitly indicates the more faithful explanation. Extensive experiments are conducted on TraP2 and two state-of-the-art approaches with relatively high accuracy.

The fidelity scores are listed in Table \ref{Fidelity_Contrastivity}. For a deeper analysis, explainers with high accuracy are not the equivalent of the interpreters with high fidelity. Although GNNExplainer exceeds Grad on all datasets under accuracy metric, it does not outperform Grad on fidelity for all cases. In contrast, Grad with lowest accuracy unexpectedly defeats the GNNExplainer in TREE-CYCLE dataset. For our method, TraP2 surpasses all alternative methods in four datasets, which proves the highest fidelity of TraP2 among them.

% It can be contributed to the perturbation in the vicinity. The explanation we extracted not only depends on a single instance but a stack of deformed graph, which results in that the selected nodes can be more robust.

\subsubsection{Question 3: Does TraP2 decisively provide explanations?}
Another valuable characteristic of an explainer is whether it can make an explanation without hesitation. As a decisive explainer, it is expected to have an ability to firmly determine a distinguished boundary among the explanation elements with high-contrast contribution value.

% Apparently, the contrastivity between the contribution of import and dispensable nodes can be an effective indicator.

We define the contrastivity by subtracting the average contribution score of the nodes outside of the $n$ nodes, extracted from Question 1, from the lowest score inside of the $n$ nodes. A higher contrastivity implies that a clear boundary is simple to identify.

As shown in Table \ref{Fidelity_Contrastivity}, TraP2 achieves the highest contrastivity value across all datasets and approaches. Especially, the superiority is apparently clear that the improvements on GNNExplainer and Grad are about 3.6 and 4.9 times on average respectively. Compared with the low contrastivity of Grad and GNNExplainer, Trap2 can be regarded as a definitely high-contrast method among them.
\begin{figure}[!t]
	%\centering
	\centerline{\includegraphics[width=1.0\textwidth]{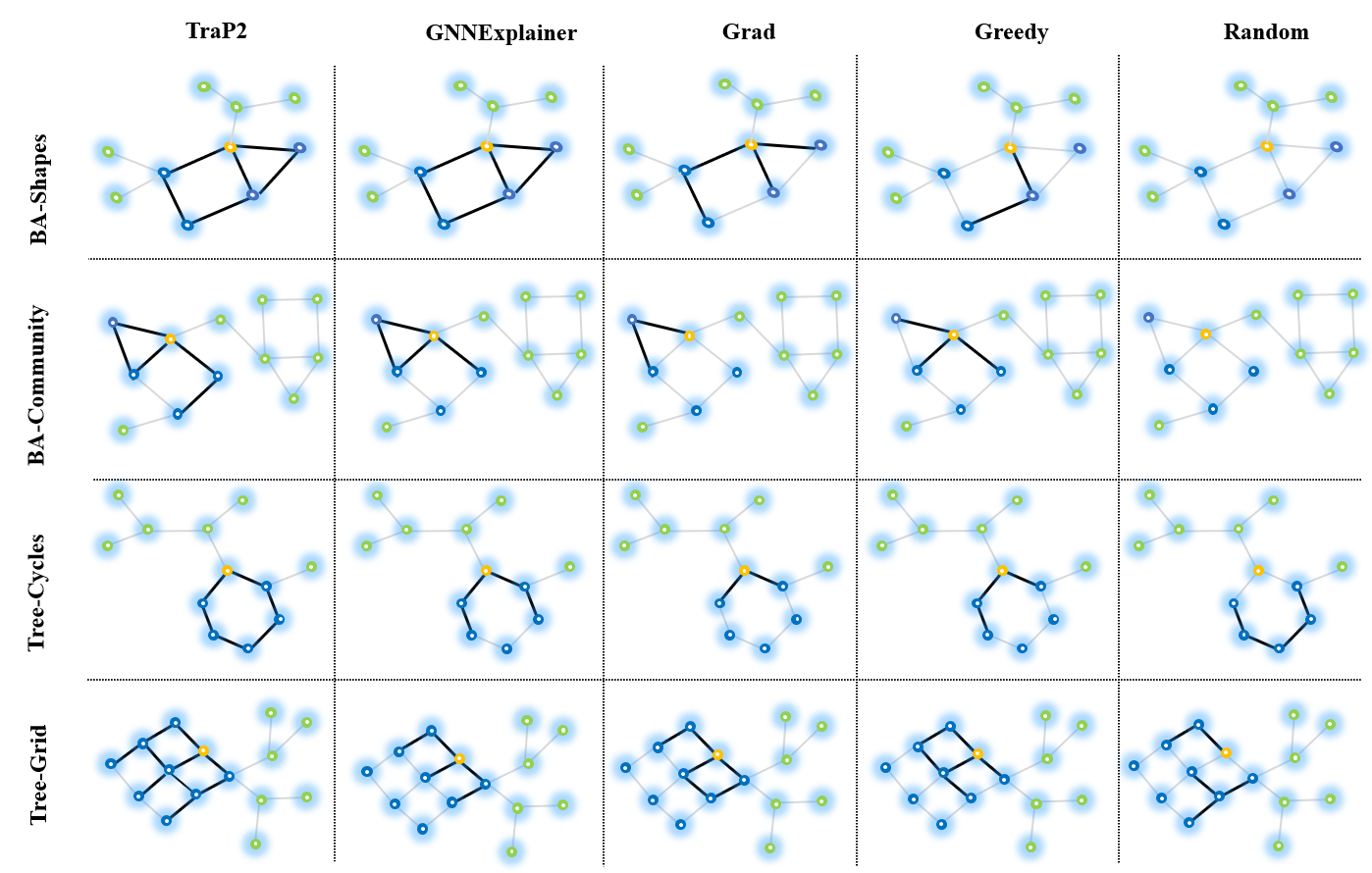}}
	\caption{Visual comparisons of the node classification examples on four datasets. The blue nodes indicate the ground truth, the interpreted node is marked in orange. Best viewed on a computer screen.}
	\label{fig:node}
\end{figure}

\subsubsection{Question 4: Can TraP2 make insightful explanation for models' predictions?}

\begin{figure}[!t]
	%\centering
	\centerline{\includegraphics[width=1.0\textwidth]{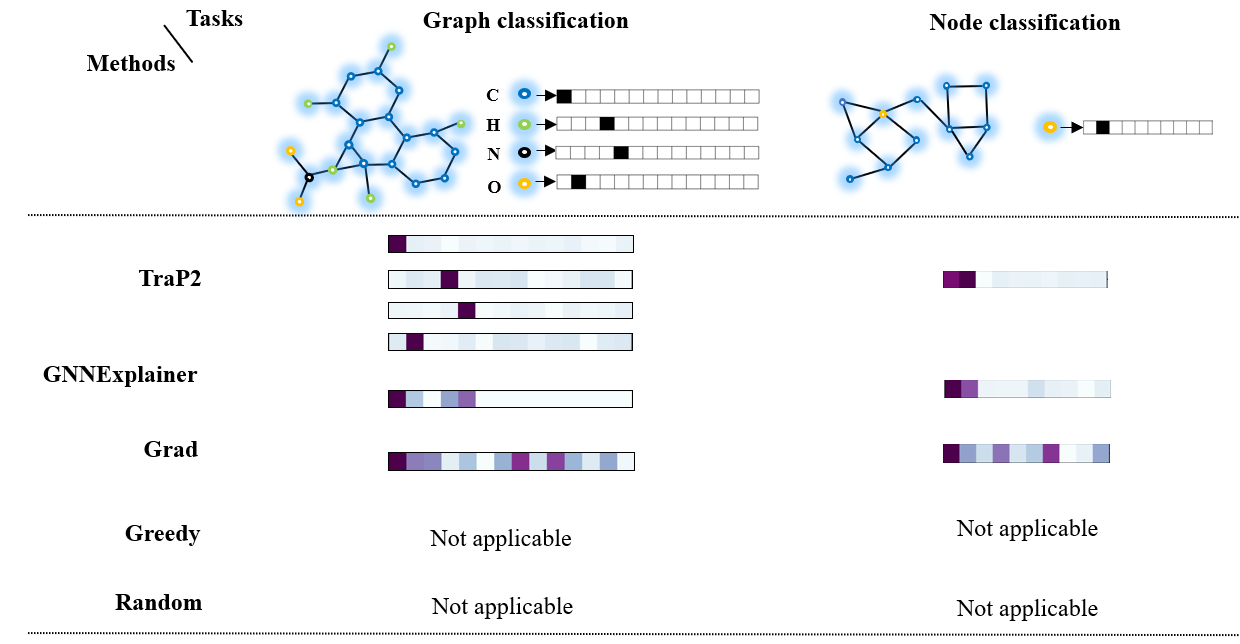}}
	\caption{Visual comparisons of the node features on MUTAG and BA-Community datasets. Best viewed on a computer screen.}
	\label{fig:node_fmap}
\end{figure}

In addition to accurately identifying the relevant subgraph structure, an explainer should highlight the most meaningful features inside the node and provide a more insightful explanation for understanding model. We compare the selected important features in each node for GNN's prediction with different approaches. In BA-Community dataset (Figure \ref{fig:node_fmap}), the TraP2 correctly recognizes the important feature component in related nodes. Furthermore, a toy experiment is conducted using same mechanism in GNNExplainer for node-independent feature contribution, which achieves the accuracy of $70.3$ and still higher than other baselines. Compared to TraP2 with node-dependent feature contribution , accuracy decreases by $0.9\%$. It implies that the joint optimization of subgraph structure and node-dependent features in our method is helpful to improve the accuracy of explanation. Similarly, in Figure \ref{fig:node_fmap}, the atoms (i.e. C, H, O and N) of molecule have different features in graph. These node features of each atom are also identified by TraP2 accurately. In comparison, GNNExplainer also captures the most important four positions of features, but does not achieve fine-grained discrimination. However, other methods can hardly identify or give incorrect explanations inside the node.

%without considering the weight of features of each node in our method is conducted,
\subsubsection{Question 5: How TraP2 find inspirational pattern from graphs?}
\begin{figure}[!t]
	%\centering
	\centerline{\includegraphics[width=1.0\textwidth]{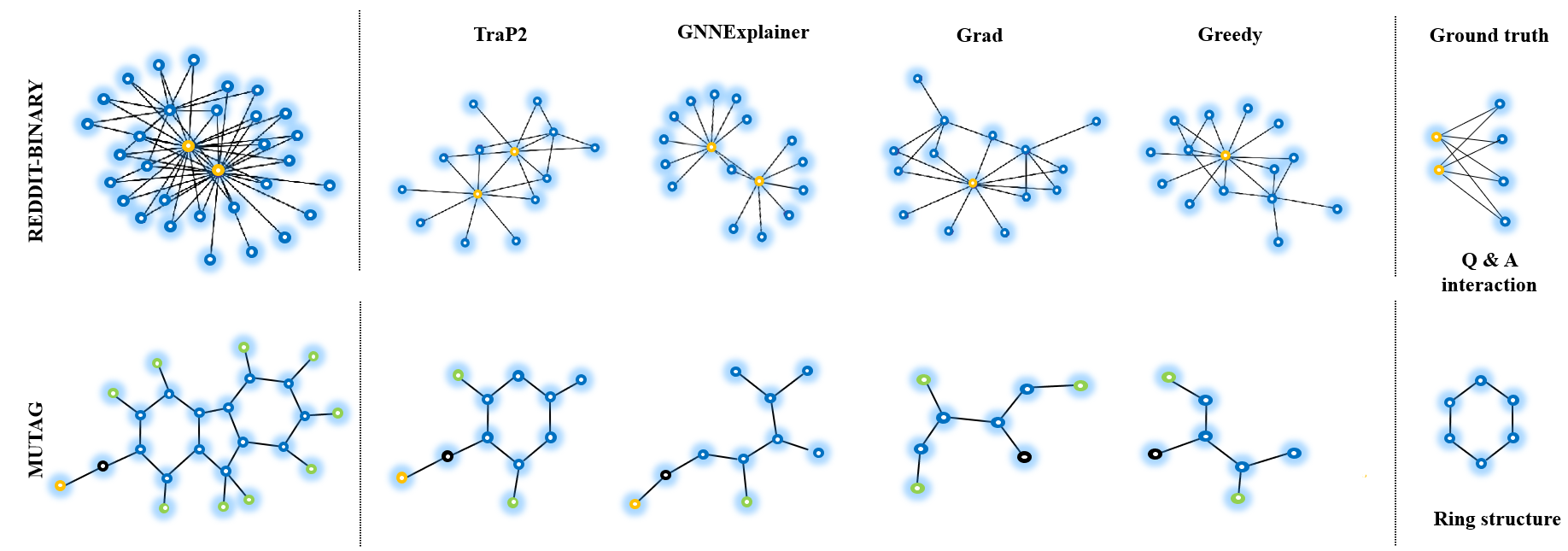}}
	\caption{Visual comparisons of the graph classification examples on two datasets. Best viewed on a computer screen.}
	\label{fig:graph}
\end{figure}
% Account for review 2 - 7
Supposing we have no prior knowledge about the GNNs - treating the original model as a black box, can the explainers provide us with a meaningful discovery? Obviously Grad requires exposed internal details of explained GNNs, while Trap2 and other baselines can be all categorized as model-agnostic approaches that no details are accessible for explainers.
To answer this question, we revisit the visualization results. For node classification, Figure \ref{fig:node} shows the different methods for four different datasets and highlights the explanation subgraphs of each method. From the figure, it can be seen that TraP2 achieves the best identification performances of key structures of house, cycle and grid. Specifically, Grad and Greedy strategy can recognize part real structures, but they are unsatisfactory because they can not provide an intuitive and understandable subgraph structure. Compared with these two models, GNNExplainer locates more crucial nodes. However, it still encounters some incomplete solutions. But the prediction effect of the random method varies greatly each time, which is reflected in both the accuracy and unstable subgraph structure. As illustrated in Figure \ref{fig:graph}, for REDDIT-BINARY dataset, the task is to identify whether a given graph belongs to a question/answer-based (Q \& A) community. TraP2 automatically finds two dense interaction patterns from complex relationship network. The original network actually represents a Q \& A interaction between two experts (the yellow nodes) and multiple visitors (the blue nodes). Similarly, the carbon ring is correctly identified by TraP2, which indicates the mutagenic factor in MUTAG dataset.

\section{Conclusion}

In this paper, based on local fidelity, we propose a novel explanation framework TraP2, in which local behaviors probed from perturbed instances in vicinity are trapped to generate a high-faithful explanation. Our explanation further explain each task on not only node but also features inside each nodes. We also propose the fidelity and the contrastivity evaluation metrics to validate the explanation performances. Extensive comparative evaluations on multiple datasets are implemented, which validates the superiority of TraP2 over several state-of-the-art explainers for GNNs. Overall, TraP2 is well adapted and applied to different interpretation tasks, which provides better explanation performance for GNNs. Based on the outstanding performance of our work, we will extend our TraP2 to support more graph mining task such as link prediction and graph generation.



\newpage
\bibliographystyle{elsarticle-num}
\bibliography{references}


% \end{thebibliography}
\end{document}